\documentclass[12pt,a4paper,twoside]{article}
\makeatletter\if@twocolumn\PassOptionsToPackage{switch}{lineno}\else\fi\makeatother

\usepackage{amsfonts,amssymb,amsbsy,latexsym,amsmath,tabulary,graphicx,times,xcolor}
\usepackage[T1]{fontenc}
\usepackage{url,multirow,morefloats,floatflt,cancel,tfrupee,multicol}
\makeatletter

\AtBeginDocument{\@ifpackageloaded{textcomp}{}{\usepackage{textcomp}}}
\makeatother
\usepackage{colortbl}
\usepackage{xcolor}
\usepackage{pifont}
\usepackage[nointegrals]{wasysym}
\usepackage[superscript,biblabel]{cite}
\usepackage{graphicx}
\makeatletter

\@ifundefined{etal}{\def\etal{\textit{et~al}}}{}

\AtBeginDocument{
\expandafter\ifx\csname eqalign\endcsname\relax
\def\eqalign#1{\null\vcenter{\def\\{\cr}\openup\jot\m@th
  \ialign{\strut$\displaystyle{##}$\hfil&$\displaystyle{{}##}$\hfil
      \crcr#1\crcr}}\,}
\fi
}

\@ifundefined{tsGraphicsScaleX}{\gdef\tsGraphicsScaleX{1}}{}
\@ifundefined{tsGraphicsScaleY}{\gdef\tsGraphicsScaleY{.9}}{}
\def\checkGraphicsWidth{\ifdim\Gin@nat@width>\linewidth
	\tsGraphicsScaleX\linewidth\else\Gin@nat@width\fi}

\def\checkGraphicsHeight{\ifdim\Gin@nat@height>.9\textheight
	\tsGraphicsScaleY\textheight\else\Gin@nat@height\fi}

\DeclareMathAlphabet{\mathpzc}{OT1}{pzc}{m}{it}

\def\URL#1#2{\@ifundefined{href}{#2}{\href{#1}{#2}}}

\edef\fntEncoding{\f@encoding}

\makeatother

\newif\ifmultipleabstract\multipleabstractfalse%
\usepackage[hmargin=1.8cm,vmargin=2.2cm]{geometry}
\makeatletter
\def\author#1{\gdef\@author{\hskip-\dimexpr(\tabcolsep)\hskip1pt\parbox{\dimexpr\textwidth-1pt}{\centering #1}}}

\let\@articletype\@empty \def\articletype#1{\gdef\@articletype{{\fontsize{14}{16}\selectfont #1}}}

\usepackage{fancyhdr}

\fancypagestyle{headings}{\fancyhf{}\fancyhead[RE]{\MakeTextUppercase{\textbf{\RunningAuthor}}}\fancyhead[LE]{\textbf{\thepage}}\fancyhead[LO]{\MakeTextUppercase{\textbf{\RunningHead}}}\fancyhead[RO]{\textbf{\thepage}}}
\fancypagestyle{plain}{\fancyhf{}\fancyfoot[C]{\textbf{\thepage}}}

\AtBeginDocument{\usepackage{lastpage}}
\def\title#1{%
  \gdef\@title{%
    \ifx\@articletype\@empty\else\@articletype~\\\fi%
     #1}%
}

\usepackage{placeins}
\usepackage{abstract}
\def\abstractname{\textbf{Abstract}}
\renewenvironment{onecolabstract}
{\vspace*{-.4pc}\trivlist\item[]\leftskip1pt\noindent\selectfont\hfill\abstractname\hfill\mbox{\null}\par\ignorespaces}{\endtrivlist}

\def\NormalBaseline{\def\baselinestretch{1.1}}

\usepackage{textcase}
\usepackage[explicit]{titlesec}
\titleformat{\section}[block]{\NormalBaseline\boldmath\bfseries}
{\thesection.}
{6pt}
{#1}
[]
\titleformat{\subsection}[hang]{\NormalBaseline\filright\itshape}
{\thesubsection.}
{6pt}
{#1}
[]
\titleformat{\subsubsection}[runin]{\NormalBaseline\filright\itshape}
{\hspace{16pt}\thesubsubsection}
{6pt}
{#1}
[]
\titleformat{\paragraph}[runin]{\NormalBaseline}
{\theparagraph}
{6pt}
{#1}
[]
\titleformat{\subparagraph}[runin]{\NormalBaseline}
{\thesubparagraph}
{6pt}
{#1}
[]

\titlespacing{\section}{0pt}{1.5\baselineskip}{.2\baselineskip}  
\titlespacing{\subsection}{0pt}{1.5\baselineskip}{.2\baselineskip}  
\titlespacing{\subsubsection}{0pt}{1.5\baselineskip}{.2\baselineskip}  
\titlespacing{\paragraph}{0pt}{.5\baselineskip}{10pt}  
\titlespacing{\subparagraph}{0pt}{.5\baselineskip}{10pt}

\usepackage{caption}
\DeclareCaptionLabelFormat{figlabel}{Figure #2}
\date{}
\makeatother

\usepackage{float}
\usepackage{booktabs}        
\usepackage{threeparttable}
\usepackage[normalem]{ulem}
\usepackage{enumitem}

\begin{document}

\title{Inchworm-Inspired Soft Robot with Groove-Guided Locomotion}
\def\RunningHead{Inchworm-inspired Multidirectional Soft Robot}
\def\RunningAuthor{H.P. Thanabalan \etal}
\author{
Hari Prakash Thanabalan\textsuperscript{1}, Lars Bengtsson\textsuperscript{1}, Ugo Lafont\textsuperscript{2}, Giovanni Volpe\textsuperscript{1} 
\thanks{\textsuperscript{1}Department of Physics, University of Gothenburg, 41296 Gothenburg, Sweden.}%
\thanks{\textsuperscript{2} European Space Research and Technology Centre (ESTEC), European Space Agency, 2201 AZ Noordwijk, Netherlands.}%
}
\maketitle


{\begin{onecolabstract}
Soft robots require directional control to navigate complex terrains. However, achieving such control often requires multiple actuators, which increases mechanical complexity, complicates control systems, and raises energy consumption. Here, we introduce an inchworm-inspired soft robot whose locomotion direction is controlled passively by patterned substrates. The robot employs a single rolled dielectric elastomer actuator, while groove patterns on a 3D-printed substrate guide its alignment and trajectory. Through systematic experiments, we demonstrate that varying groove angles enables precise control of locomotion direction without the need for complex actuation strategies. This groove-guided approach reduces energy consumption, simplifies robot design, and expands the applicability of bio-inspired soft robots in fields such as search and rescue, pipe inspection, and planetary exploration.

\def\keywordstitle{Keywords}
\end{onecolabstract}}
 
\begin{multicols}{2}

\section*{Introduction}

Bioinspired and biomimetic design principles have always fascinated soft roboticists because of Nature's remarkable adaptability to complex environments. Many soft robot designs take inspiration from biological systems, such as snakes \cite{VirgalaSnake}, fish \cite{doi:10.1089/soro.2017.0062}, jellyfish \cite{godaba2016soft}, and even plants \cite{8487848}. Inchworms and earthworms have been of particular interest to soft roboticists. These organisms have evolved distinctive locomotion strategies that allow them to traverse dry and unstructured terrain using simple, yet effective body structures. Whereas earthworms move in a peristaltic manner \cite{niu2015enabling}, inchworms locomote through sequential anchoring and stretching \cite{PLAUT201556}: the prolegs adhere to the surface, the body elongates forward, and contraction of anterior muscles pulls the head and thorax ahead. This simple yet effective strategy enables them to navigate diverse and complex environments.

However, for soft robots to move in complex terrains, directional control is critical for adaptability. Despite significant progress in soft robotics, achieving multidirectional locomotion remains challenging. Soft robots rely on compliant actuators and structures, making it difficult to generate controllable motion in multiple directions. As a result, soft roboticists often employ multiple actuators \cite{10449469, doi:10.1126/scirobotics.aaz6451, CAO20189, 8404936, LI2023114412, https://doi.org/10.1002/adfm.202305046}, each contributing to specific deformation patterns that enable directional motion. This approach, however, increases mechanical and control complexity as well as energy consumption.

\begin{figure*}[!htbp]
    \includegraphics[width=\textwidth]{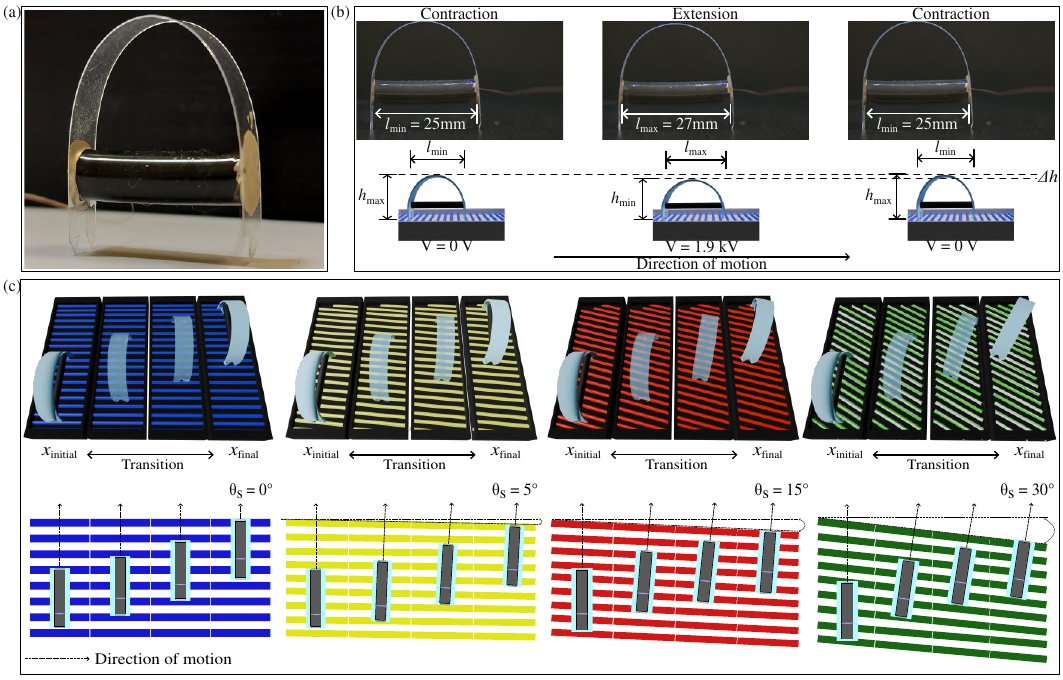}
    \caption{
    {\bf Geometry, actuation cycle, and groove-guided orientation of the inchworm-inspired soft robot.} 
    ({\bf a}) Photograph of the soft robot, consisting of a multilayer rolled dielectric elastomer actuator (RDEA) integrated with a flexible PET sheet. 
    ({\bf b}) Actuation cycle during locomotion: the robot contracts at 0~V with a minimum length $l_{\min}=25$~mm and maximum height $h_{\max}$; upon applying 1.9~kV, the actuator extends to $l_{\max}=27$~mm with reduced height $h_{\min}$; returning to 0~V restores the contracted state (see {Supplementary Movie S1}). This contraction--extension cycle drives the robot forward along the substrate. 
    ({\bf c}) Schematic of the robot on 3D-printed grooved substrates at different groove orientations ($\theta_{\rm s}=0^\circ,\,5^\circ,\,15^\circ,\,30^\circ$). The groove angles bias frictional engagement, passively steering the robot and aligning its direction of motion with the groove orientation.
    }
    \label{FigureLabel1}
\end{figure*}

The actuators employed in soft robotics include pneumatic \cite{10585360}, magnetic \cite{doi:10.1089/soro.2018.0082}, light \cite{https://doi.org/10.1002/aisy.202300466}, shape memory alloy (SMA) \cite{10.1007/978-981-97-3530-3_28}, ionic polymer-metal composites \cite{inproceedings}, and dielectric elastomer actuators (DEAs) \cite{10.1117/12.2044573}. Amongst these, DEAs are particularly promising because they exhibit muscle-like behaviour, with large deformation, high energy density, high elastic modulus (on the order of 1~MPa), fast response times (typically under 1~ms), and high efficiency \cite{godaba2016soft, shintake2018soft,hajiesmaili2021dielectric,WANG2022308}.
DEAs have been employed in a wide range of soft robotics applications, including insect-inspired robots fabricated via multimaterial 3D printing \cite{zhu20233d}, inchworm robots with integrated feedback control \cite{cao2019control}, snap-buckling balloon actuators inspired by the Venus flytrap \cite{baumgartner2020lesson}, and suction mechanisms modelled after octopus tentacles \cite{jamali2024soft}.

Several studies have shown the effectiveness of using DEAs as soft actuators for developing inchworm-inspired soft robots.
Duduta et al. \cite{7989501} reported a multi-layered, high-speed DEA (one body length per second) inchworm with high responsive time, reduced actuation voltage, and high energy density \cite{hajiesmaili2021dielectric}.
Hu et al. \cite{https://doi.org/10.1002/aisy.202200209} used a DEA constrained to acrylic sticks attached to a flexible support frame with two electro-adhesive actuators acting as anchors, enabling locomotion on both horizontal and vertical surfaces. 
Li et al. \cite{doi:10.1089/soro.2018.0037} used a rolled DEA (RDEA) wrapped around a compressed spring to mimic inchworm locomotion. 
Xu et al. \cite{xu2017bio} developed a multi-segmented earthworm-inspired robot by connecting multiple DEAs with bristles attached at the bottom to generate anisotropic friction, imitating biological setae.
Liu et al. \cite{liu2021bioinspired} fabricated an inchworm-inspired triboelectric soft robot powered by a rotary freestanding triboelectric nanogenerator, capable of locomotion at 14.9~mm/s under 5~kV and equipped with triboelectric adhesion feet for movement across diverse surfaces.

Despite these advances, only a handful of DEA-based soft robots are capable of multidirectionality.
Digumarti et al. \cite{8404936} fabricated a multidirectional soft robot that has a four-sector disk, where each sector consists of a single DEA that can be actuated independently.
Guo et al. \cite{GUO2022101720} fabricated a multidirectional soft robot using a two-degrees-of-freedom DEA embedded within a single RDEA and two flexible electroadhesives.
Wang et al. \cite{wang2023dexterous} designed a dexterous robot that is capable of multidirectional locomotion by using a reconfigurable chiral-lattice foot design and a single DEA. The robot could achieve multidirectionality by three methods: First, by manually twisting the chiral-lattice foot, clockwise for left turn and counter-clockwise for right turn; second by heating and cooling of the chiral-lattice foot which compresses it and induces turning; and third by changing the frequency of the actuator for forward and backward movement. However, this system required very high voltages (6–8~kV) and manual intervention, limiting its practicality.

In this study, we present a minimalistic yet effective approach for achieving multidirectional locomotion without manual intervention.
As shown in Figs.~1{\bf a} and 1{\bf b}, the robot uses a single multilayered RDEA coupled with a PET sheet, while groove-guided passive alignment on 3D-printed substrates provides a compact and scalable platform for directional control. The key innovation lies in the incorporation of strategically designed groove patterns on the substrate. By varying their orientation, the robot can passively align its body and adjust its trajectory, thereby achieving autonomous multidirectional locomotion (Fig.~1{\bf c}).

The remainder of this paper is organised as follows: Section~2 (Materials and Methods) describes the actuator configuration, groove design parameters, and assembly process. Section~3 (Results) presents an in-depth analysis of locomotion performance across different substrate configurations and orientations, demonstrating the effectiveness of groove-guided control. Section~4 (Discussion and Conclusion) summarises the key findings and outlines future research directions.

\section*{Materials and Methods}

\begin{figure*}[!htbp]
    \centering
    \includegraphics[width=\textwidth]{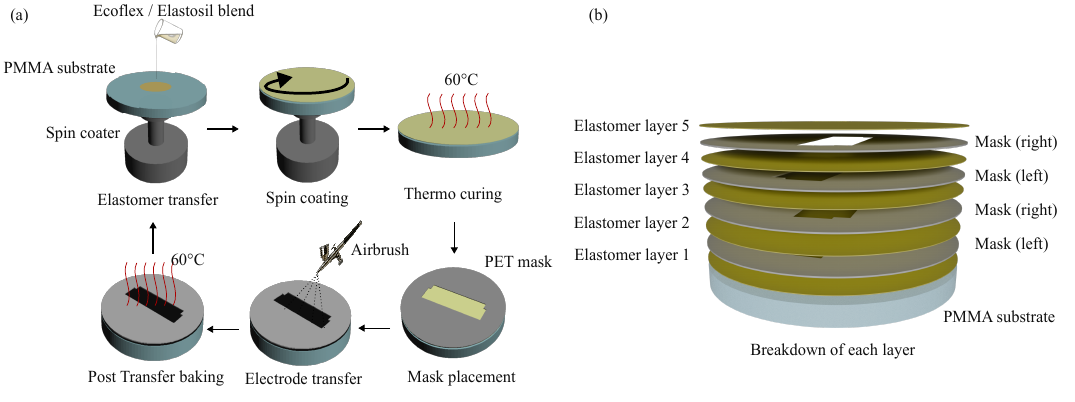}
    \caption{
    {\bf Fabrication process of the rolled dielectric elastomer actuator (RDEA).} 
    ({\bf a}) Step-by-step illustration of the fabrication workflow. The Ecoflex/Elastosil blend is first deposited onto a PMMA substrate and uniformly spin coated. The coated film is thermally cured at 60\textdegree C. A PET mask is then aligned on the cured layer to define the electrode regions, and compliant electrodes are deposited using an airbrush. The sample is subsequently baked at 60\textdegree C to remove excess solvent (post-transfer baking). These steps (spin coating, curing, mask placement, electrode transfer, and baking) are repeated to build a multilayer actuator. 
    ({\bf b}) Layer-by-layer breakdown of the final 5-layer RDEA structure. Alternating elastomer layers and electrode masks (left and right configurations) are stacked to ensure electrical connectivity to opposite terminals.
    }
    \label{FigureLabel2}
\end{figure*}

\begin{table*}
\centering
\begin{threeparttable}
\caption{Fabrication parameters for 5-layer RDEA}
\label{tab:RDEA_fabrication}
\setlength{\tabcolsep}{6pt} 
\begin{tabular}{@{}c c c c c@{}} 
\toprule
Layer & Spin coat speed (rpm)\tnote{$a$} & Vacuum (min)\tnote{$b$} & Thermal curing (min)\tnote{$c$} & Post-transfer bake (min)\tnote{$d$} \\  
\midrule
1 & 2100 & 3.0 & 3.0 & 3.0 \\ 
2 & 1500 & 3.0 & 4.0 & 3.0 \\ 
3 & 1500 & 3.0 & 4.0 & 3.0 \\ 
4 & 1500 & 3.0 & 4.5 & 3.0 \\ 
5 & 2400 & 3.0 & 5.5 & 3.0 \\ 
\bottomrule
\end{tabular}
\begin{tablenotes}[flushleft]\footnotesize
\item[${a}$] Spin coat speed refers to the rotation speed of the PMMA substrate during deposition.
\item[${b}$] Vacuum indicates the time under vacuum before curing.
\item[${c,d}$]Thermal curing and post-transfer bake durations are at 60\textdegree C.
\end{tablenotes}
\end{threeparttable}
\end{table*}

\subsection*{Rolled Dielectric Elastomer Actuator (RDEA) Fabrication}

Two silicone elastomers were selected as the dielectric layers of the elastomer matrix. The first was Ecoflex 00-30 (Smooth-On Inc., Macungie, PA, USA), an ultra-soft, bi-component, platinum-catalysed elastomer that cures at room temperature. The second was Elastosil P7670 (Wacker Chemie AG, Germany), chosen to complement Ecoflex in the composite matrix. The reasons for selecting these two materials are discussed in detail below when discussing the fabrication process (Step (2) RDEA fabrication).

For the compliant electrodes, single walled carbon nanotubes (P3-SWCNTs, Carbon Solution Inc.) with an average particle diameter of 2–5~nm were employed. Isopropyl alcohol ($\geq$~99.5\%, Sigma-Aldrich) and deionised water were used as solvents for dispersing the carbon nanotube particles. Electrical connections were completed with a two-part (Part A epoxy and Part B hardener) conductive silver epoxy (60 Minute Conductive Silver Epoxy, CW2460, Chemtronics, USA).

The fabrication process was divided into two parts: (1) Compliant electrode preparation, and (2) RDEA fabrication.

For step (1), the compliant electrode was prepared. First, 7~mg of P3-SWCNT were dispersed in 2~ml of deionised water and 20~ml of isopropyl alcohol. Then, this dispersion was sonicated in an ultrasonic ice bath for two hours and then centrifuged at 4,400 r.p.m. for one hour to achieve a homogeneous suspension and remove any aggregates from dispersions. 

For step (2), the fabrication method of the RDEA was adapted from Ref.~\citeonline{https://doi.org/10.1002/adma.202106757}, and it is illustrated in Fig.~2{\bf a}. A circular acrylic substrate (PMMA) with 16~cm diameter was employed as supporting platform. The elastomeric layer was prepared using a blend of 70\% Elastosil P7670 and 30 \% Ecoflex 0030. This specific composition of the Elastosil/Ecoflex composite was formulated to achieve enhanced dielectric strength, improved ageing and tear resistance, as well as superior durability.  While 100\% Elastosil P7670 provides similar levels of dielectric strength and durability, the Elastosil/Ecoflex blend offers significantly higher stretchability due to the high elongation at break percentage of Ecoflex 00-30. Conversely, 100\% Ecoflex 00-30 exhibits substantially higher stretchability, but it has a lower dielectric strength, lower tear resistance, and lower durability. First, the silicone mixture was manually mixed for a few minutes before being placed in a vacuum chamber (KnF Neuberger, Germany) for ten minutes for degassing. Once degassed, the silicone--elastomer blend was carefully poured onto the PMMA substrate for spin coating. After the first layer had been spin-coated, the substrate was placed in a vacuum chamber for three minutes to further remove any remaining air at 60\textdegree C for curing.

Once the silicone elastomer was cured and removed from the oven, a PET sheet with the electrode pattern (mask) was placed on the silicone elastomer. Here, two types of masks were prepared and applied alternately for each layer, enabling the compliant electrode in the odd and even layers to connect to opposite electrical terminals. To ensure accuracy and clean cutting of the mask, a laser cutter was employed to cut the electrode pattern onto a PET sheet. Then, using a syringe, the electrode solution was extracted and placed into the fluid cup of the airbrush. The flow rate was kept constant at 1~ml/min at 200~kPa to ensure consistent electrode deposition. While in Ref.~\citeonline{https://doi.org/10.1002/adma.202106757} a stamping method was applied, we used a spraying approach with an airbrush as it is able to cover a much larger contact area. Finally, after the electrode transfer, the substrate was cured at 60\textdegree C (post baking transfer) for three minutes to allow excess solvent to evaporate. This process was then repeated for five layers of dielectric membranes and four layers of compliant electrodes. The fabrication parameters for this 5-layer RDEA are reported in Table~1 and the breakdown of the layers is shown in Fig.~2{\bf b}.

\newpage
\subsection*{Inchworm Soft Robot Fabrication}

\begin{figure*}[hb!]
    \centering
    \includegraphics[width=1\textwidth]{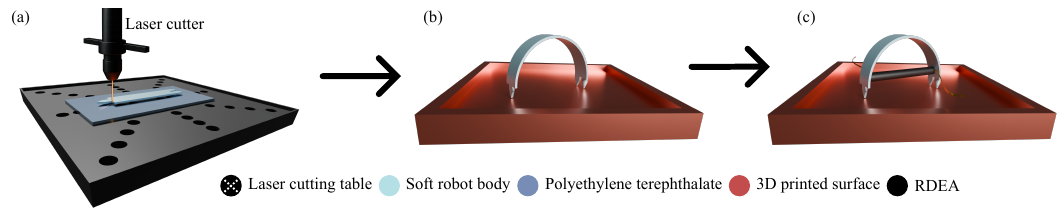}
    \caption{
    {\bf Inchworm soft robot fabrication.} ({\bf a}) Using a laser cutter, a flexible polyethylene terephthalate (PET) sheet was used to fabricate the body of the soft robot. ({\bf b}) The PET was bent at the edges to form an arc. ({\bf c}) The RDEA was placed in between the arc and was secured using conductive epoxy.  
    }
    \label{FigureLabel3}
\end{figure*}

To keep the soft robot design minimal, the body was fabricated using commercially available flexible PET sheets as shown in Fig.~3. Laser cutting technology was employed to precisely cut the robot body, ensuring clean edges (Fig.~3{\bf a}). Once the body had been cut, the PET sheet was bent downward to form an arc (Fig.~3{\bf b}). The RDEA was then positioned between the bent PET sections, which were glued to the corresponding sides of the RDEA using conductive epoxy, while copper wires were attached to each side (Fig.~3{\bf c}). The conductive epoxy was left to cure at room temperature for about an hour.

Actuation was driven using a waveform generator and a high-voltage power supply at a maximum voltage of 1.9~kV (PS350/5000V-25W, Stanford Research Systems Inc.) and a maximum frequency of 400~mHz (T3AFG30, Teledyne, 30 MHz). This frequency was determined as optimal for locomotion, as it enabled the front leg to reliably engage with each groove and maintain effective grip during the actuation cycle. At higher frequencies, the front leg tended to slide and had a poor grip on the grooves, preventing directional motion, whereas at lower frequencies the robot achieved lower speed. 

\subsection*{3D-Printed Substrate Fabrication}

To enable directional locomotion and allow the robot to move at various angles, a custom 3D-printed substrate was designed and fabricated. The design process began with creating a 3D model of the substrate in CAD software (FreeCAD), where the overall dimensions  and surface features were defined. The model was then exported in STL format and imported into the 3D printer’s slicing software (FlashPrint 5). 

Before printing, several parameters were adjusted to ensure high quality fabrication. These included setting the appropriate layer height, infill density, print speed, and support structures, as well as optimising the bed and extruder temperatures for the chosen printing material which was Polylactic Acid (PLA). 

Substrates were fabricated with varying groove angles (0°, 5°, 15°, and 30°). This was achieved by re-orienting the CAD model within the slicing software prior to printing. Once the parameters were finalised, the prepared G-code (GCODE) file was transferred to the printer (FlashForge Adventure 4).

After fabrication, the substrates were carefully removed from the build plate, inspected for surface uniformity, and lightly cleaned to remove any residual or filament artefacts. The grooves on the substrate appeared naturally during the 3D printing process as the PLA filament was deposited onto the printing bed. Since the extruded filament tends to solidify with a slightly curved cross-section, where this leaves behind ridge-like patterns on the surface. 

\section*{Results}

\subsection*{Effect of Substrate Groove Angle on Directional Locomotion}

Multidirectional soft robots have gained considerable attention in the field of soft robotics \cite{}. Most existing designs, however, require multiple actuators to achieve movements with more than one degree of freedom, which increases complexity \cite{8404936, GUO2022101720, wang2023dexterous}. In contrast, this work introduces a minimalistic strategy in which multidirectionality is realised with a single RDEA. To investigate this capability, we examined how varying the groove angle of the substrate influences the robot’s locomotion and orientation. The grooves act as directional friction cues, engaging with the robot’s legs to bias movement along specific paths. By systematically adjusting the groove angle from 0° to 5°, 10°, and 30°, we analysed how these geometrical changes affect the robot’s ability to reorient and maintain controlled multidirectional motion.

\subsection*{Baseline Locomotion on 0° Grooves}

To analyse the interaction between groove angle and robot orientation, the robot was first tested on a substrate with a 0° groove angle, where the surface patterns are oriented perpendicular to the robot’s axis of motion. On this substrate, the interaction between the robot’s legs and the grooves are symmetrical, with each leg experiencing comparable frictional resistance. This eliminates any differential lateral force and allows the robot to move in a straight path without unintended turning (see {Supplementary Movie S2}). Fig.~4 illustrates the robot’s locomotion on the 0° substrate. Over time, the trajectory shows negligible angular deviation, confirming that in the absence of directional surface cues, the robot maintains a straight path with no preferential reorientation.

\begin{center}
\begin{minipage}{\linewidth}
    \centering
    \includegraphics[width=\linewidth]{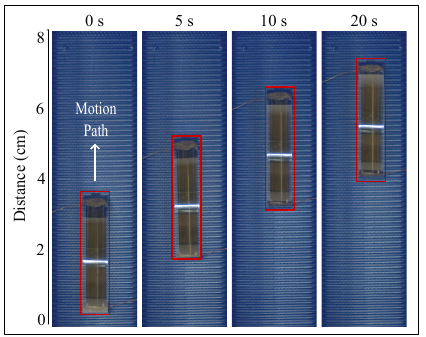}
    \captionof{figure}{
    {\bf Temporal evolution of the robot’s orientation angle on a 0° grooved substrate.}
    The robot's orientation remains nearly constant, indicating that without directional surface cues the robot follows a straight trajectory with no reorientation.
    }
    \label{FigureLabel4}
\end{minipage}
\end{center}

\subsection*{Locomotion on 5° Grooves}

Following the baseline test at 0°, the robot was evaluated on a substrate with a 5° groove angle, where the surface patterns are slightly tilted relative to the robot’s initial axis of motion. In contrast to the symmetrical interaction observed at 0°, which produced straight-line locomotion, the 5° grooves introduced an asymmetry in the frictional forces acting on the robot’s legs. This asymmetry generated a lateral bias, gradually steering the robot away from its initial orientation (see {Supplementary Movie S3}). Fig.~5 shows the temporal evolution of the robot’s orientation on the 5° substrate.

\begin{center}
\begin{minipage}{\linewidth}
    \centering
    \includegraphics[width=\linewidth]{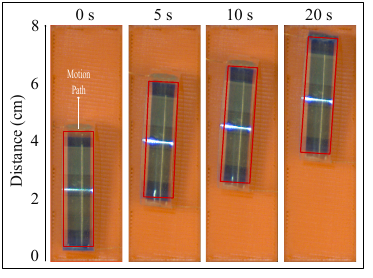}
    \captionof{figure}{
    {\bf Temporal evolution of the robot’s orientation angle on a 5° grooved substrate.}
    The robot exhibits small, but systematic angular deviations, demonstrating that shallow surface cues effectively guide locomotion and induce gradual reorientation consistent with the groove angle.
    }
    \label{FigureLabel5}
\end{minipage}
\end{center}

Over time, a slight but consistent angular deviation appears, indicating that even shallow directional cues influence the robot’s trajectory in accordance with the 5° groove orientation.

\subsection*{Locomotion on 15° and 30° Grooves}

Following the test on a 5° substrate, the groove angle was increased to 15° and 30° to observe the robot’s behaviour under more pronounced directional cues. At 15° (Fig.~6, top), the robot exhibits a clearer reorientation compared to the 5° case, indicating stronger guidance by the substrate (see {Supplementary Movie S4}).

\begin{center}
\begin{minipage}{\linewidth}
    \centering
    \includegraphics[width=\linewidth]{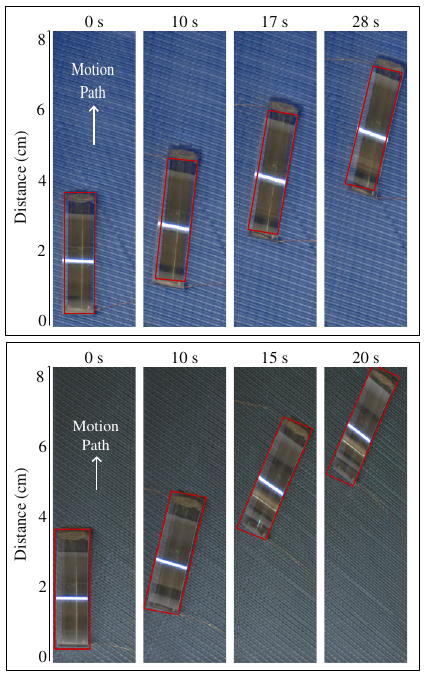} 
    \captionof{figure}{
    {\bf Temporal evolution of the robot’s orientation angle on 15° and 30° grooved substrates.}
    Increasing the groove angle produces progressively stronger alignment with the substrate orientation.
    On the 15° substrate (top), the robot undergoes moderate angular reorientation, showing that even moderate surface anisotropy effectively guides locomotion.
    On the 30° substrate (bottom), the robot consistently realigns with the groove direction, demonstrating robust directional control due to the steeper cues.
    }
    \label{FigureLabel6} 
\end{minipage}
\end{center}

At 30° (Fig.~6, bottom), the reorientation becomes much more pronounced. The effect is caused by asymmetric frictional engagement: as the groove angle increases, this asymmetry strengthens, causing the actuator-generated forces to resolve into two components: one along the groove axis and the other perpendicular to it—thereby biasing the robot’s trajectory (see {Supplementary Movie S5}). Steeper groove angles therefore introduce larger lateral force components, producing stronger turning effects that curve the locomotion path. 

\begin{figure*}[!htbp]
    \centering
    \includegraphics[width=1\textwidth]{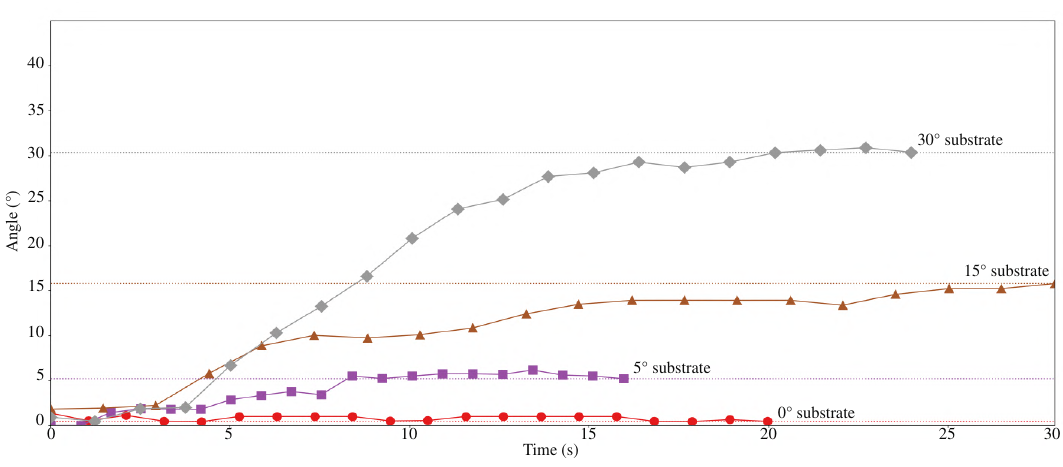}
    \caption{
    {\bf Temporal evolution of the robot’s orientation across different substrate groove angles.}  
    The y-axis shows the orientation angle of the robot, while the x-axis represents elapsed time.  
    Each curve corresponds to a different substrate angle (0°, 5°, 15°, 30°), illustrating how steeper grooves induce progressively larger reorientation toward the groove direction.
    }
    \label{FigureLabel7}
\end{figure*}

Figure~7 summarises these results, showing the temporal evolution of orientation across substrates with groove angles from 0° to 30°. The plot reveals a clear trend: while the robot maintains a straight trajectory on the 0° substrate, progressively steeper grooves at 5°, 15°, and 30° induce stronger reorientation, highlighting the effectiveness of groove geometry as a passive mechanism for guiding locomotion.

\subsection*{Trajectory Reorientation at Substrate Interfaces}

Moving beyond single substrates, the soft robot demonstrated multidirectionality by traversing multiple substrates with different groove angles. At these interfaces, the robot transitioned between substrates of varying orientations, adapting its locomotion to the changing surface geometry. 

As an initial test of adaptability, we examined locomotion across a transition between two substrates with differing groove angles. Figure~8 presents two representative cases: In Fig.~8{\bf a}, the robot moves from a 0° substrate into a +15° substrate (straight to right turn) (see {Supplementary Movie S6}). In Fig.~8{\bf b}, it moves from a 0° substrate into a –10° substrate (straight to left turn) (see {Supplementary Movie S7}). (Groove angles are defined as positive or negative to indicate turning direction: positive values correspond to right turns, while negative values correspond to left turns.)
On the initial 0° surface, the robot moved in a straight line due to symmetrical frictional interaction. Upon entering the second substrate, the front legs experienced asymmetric frictional forces induced by the groove orientation, which passively redirected the trajectory. The turning direction and magnitude were directly determined by the groove angle, aligning the robot’s path with the substrate geometry.

\begin{figure*}[!b]
    \centering
    \includegraphics[width=1.0\textwidth]{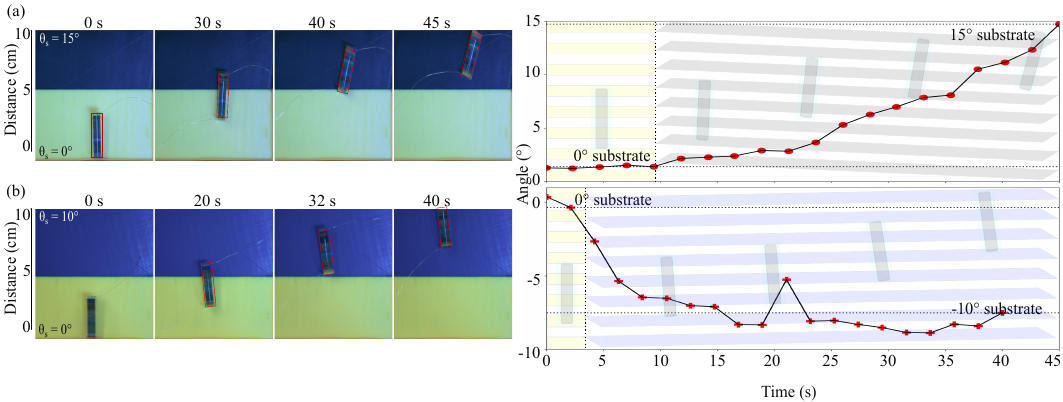}
    \caption{
    {\bf Robot trajectories across substrate interfaces.}
    ({\bf a}) Transition from a flat 0° substrate to a +15° grooved substrate (straight to right turn).
    ({\bf b}) Transition from a flat 0° substrate to a –10° grooved substrate (straight to left turn).
    In both cases, the robot moves in a straight line on the 0° region and exhibits a clear angular deviation upon entering the grooved section. The observed reorientation reflects the influence of groove angle on passive steering, with annotated angles showing the initial and final orientations.}
    \label{FigureLabel8} 
\end{figure*}

\subsection*{Trajectory Reorientation Across Multiple Substrate Interfaces}

\begin{figure*}[!hb]
    \centering
    \includegraphics[width=1.0\textwidth]{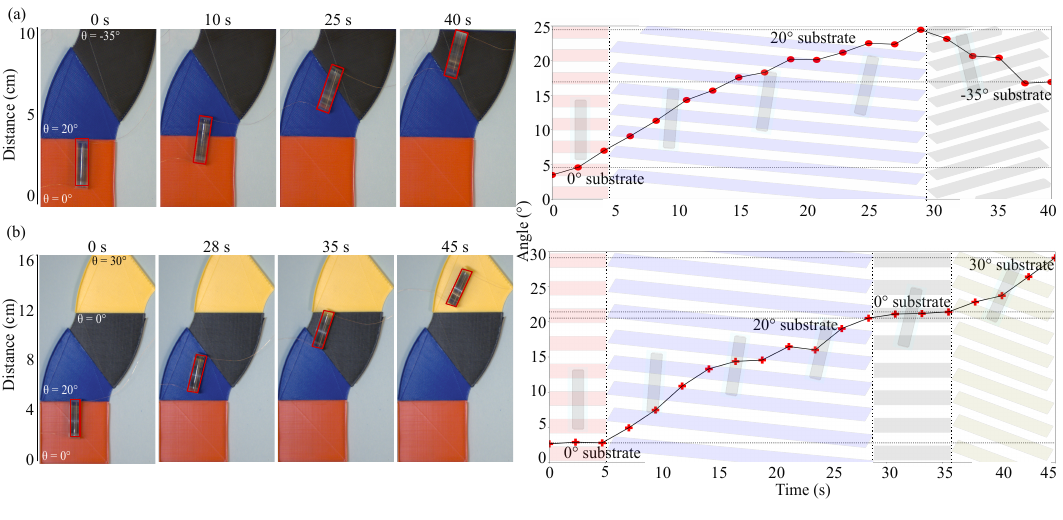}
    \caption{
    {\bf Temporal evolution of the robot’s orientation angle across multiple substrate interfaces.}
    ({\bf a}) Three-substrate configuration: starting on a 0° substrate, the robot gradually turns right on the +20° section, then shifts left on the –35° section, producing a net leftward reorientation of about –15°. Inset images showing robot positions at representative tilt angles. 
    ({\bf b}) Four-substrate configuration: starting at 0°, the robot turns right on the +20° section, maintains this trajectory across the intermediate 0° substrate, and then undergoes further rightward turning on the +30° section. Inset images showing robot positions at representative tilt angles
    }
    \label{FigureLabel9} 
\end{figure*}

Building on the dual-substrate experiments, we next evaluated the robot’s adaptability across more complex configurations involving three or four sequential substrates. These tests assessed whether the robot could maintain controlled reorientation while negotiating multiple successive transitions in surface geometry. The groove angles were arranged with both positive and negative orientations, requiring the robot to execute successive left and right turns.

In Fig.~9{\bf a}, the groove orientations are arranged at 0°, +20°, and –35°. The robot begins on the 0° substrate, moving straight due to symmetrical frictional forces. On the second substrate, the +20° groove induces a rightward turn through asymmetric frictional engagement. Upon transitioning to the –35° substrate, the robot turns left; however, the prior right turn reduces the net turning effect. The resulting trajectory corresponds to an overall leftward reorientation of approximately –15° (see {Supplementary Movie S8}). This demonstrates that the robot’s directional behaviour depends not only on the current groove angle but also on the orientation history of preceding substrates.

Similarly, in Fig.~9{\bf b}, the groove orientations are 0°, +20°, 0°, and +30°. After a rightward turn on the +20° substrate, the robot maintains its trajectory across the intermediate 0° surface. On entering the final +30° substrate, the turning effect is reinforced, producing an even greater rightward reorientation (see {Supplementary Movie S9}).

\section*{Discussion and Conclusion}  

This work presents a novel inchworm-inspired soft robot capable of passive directional control through interaction with grooved substrates. By exploiting frictional asymmetry introduced by varying groove angles, the robot achieves multidirectional locomotion and demonstrates the ability to adapt its trajectory when transitioning across substrates with different groove orientations. These results highlight how surface topology can be used as an effective passive guidance mechanism, eliminating the need for complex control architectures, multiple actuators, or manual intervention.  

A key finding is that the robot adapts not only to single substrate geometries but also to sequential transitions, where its trajectory is shaped by the cumulative mechanical influence of multiple groove interfaces. This history-dependent behaviour suggests that locomotion paths can be programmed through the deliberate arrangement of environmental features, enabling complex navigation in artificially structured terrains without active sensing.  

In addition to substrate-guided steering, we tested two different compliant electrode materials (Vulcan XC72 and P3-SWCNT) to determine whether there are any significant changes on the motion of the robot and found no significant performance differences in locomotion. This finding indicates that different compliant electrode usage could be employed without compromising locomotion efficiency, providing flexibility in material selection for applications where conductivity, durability, or cost may be critical factors. This robustness to electrode composition indicates that the locomotion mechanism is primarily governed by substrate interaction and actuator configuration rather than electrode type, which broadens material flexibility for future designs. However, the use of SWCNTs is associated with the fault-tolerant properties of the nanotubes \cite{yuan2008fault, stoyanov2013long, yuan2009dielectric}, which can make the robot more resilient to damage and potentially extend its operational lifespan.

Compared to existing approaches that rely on multiple actuators or sophisticated reconfigurable mechanisms \cite{10449469, doi:10.1126/scirobotics.aaz6451, 8404936, wang2023dexterous}, our method offers notable advantages in simplicity, scalability, and energy efficiency. While other DEA-based robots have achieved impressive locomotion, they often require high voltages, anchoring systems, or hybrid actuation strategies that increase fabrication and control complexity \cite{liu2021bioinspired, GUO2022101720}. In contrast, the groove-guided approach makes use of environmental features to reduce the burden on the robot itself while maintaining relatively low operating voltages compared to other multi-actuator systems.

Nevertheless, there are limitations where the locomotion behaviour observed reveals that the robot’s movement is strongly influenced by the structural characteristics of the substrate. In particular, when surface features such as grooves with directional textures are present, the robot consistently tends to move perpendicular to these features, indicating that its navigation becomes environmentally guided rather than purely actuation-driven. As a result, on randomly patterned terrains, the robot may follow preferential paths dictated by the substrate, which can facilitate passive exploration but simultaneously might limit precise directional control. In addition, surface hardness might play a role and shall be investigated in view of having such robot operation on natural terrain. Moreover, although the RDEA achieved reliable actuation at 1.9~kV, further reduction of operating voltage would be essential for practical deployment. Addressing these challenges could involve integrating passive groove-guidance with onboard sensing or hybrid control strategies, thereby combining environmental simplicity with active adaptability.  

Looking ahead, this approach opens up opportunities for soft robots designed to operate in heterogeneous environments where minimal sensing and low onboard complexity are critical, such as search and rescue, planetary exploration, and in-situ inspection. Extending groove-guided locomotion to irregular or stochastic surface patterns, and investigating scaling effects for micro- and macroscale robots, represent promising directions for future research.

\section*{Acknowledgements}

This work was performed under the European Space Agency (Contract No. 4000139030).
Furthermore, the authors acknowledge the support from the Horizon Europe ERC Consolidator Grant MAPEI (Grant No.
101001267),the Knut and Alice Wallenberg Foundation (Grant No. 2019.0079), and Adlerbert Research Foundation.

\section*{Author Disclosure Statement}

The authors have no competing interests.

\bibliographystyle{ieeetr}
\bibliography{references.bib}

\end{multicols}
\end{document}